\pdfoutput=1
\documentclass[letterpaper]{article} % DO NOT CHANGE THIS
\usepackage{aaai24}  % DO NOT CHANGE THIS
\usepackage{times}  % DO NOT CHANGE THIS
\usepackage{helvet}  % DO NOT CHANGE THIS
\usepackage{courier}  % DO NOT CHANGE THIS
\usepackage[hyphens]{url}  % DO NOT CHANGE THIS
\usepackage{graphicx} % DO NOT CHANGE THIS
\usepackage{natbib}  % DO NOT CHANGE THIS AND DO NOT ADD ANY OPTIONS TO IT
\usepackage{caption} % DO NOT CHANGE THIS AND DO NOT ADD ANY OPTIONS TO IT
\usepackage{algorithm}
\usepackage{algorithmic}

\usepackage{newfloat}
\usepackage{listings}
\DeclareCaptionStyle{ruled}{labelfont=normalfont,labelsep=colon,strut=off} % DO NOT CHANGE THIS
\floatstyle{ruled}
\newfloat{listing}{tb}{lst}{}
\floatname{listing}{Listing}
\usepackage{utfsym}
\usepackage{booktabs}
\usepackage{multirow}
\usepackage{pifont}
\usepackage{makecell}
\usepackage{amsmath}

\newcommand{\Tref}[1]{Table~\ref{#1}}
\newcommand{\Eref}[1]{Equation~(\ref{#1})}
\newcommand{\Fref}[1]{Figure~\ref{#1}}

\newcommand{\eg}{\textit{e.g.}}
\newcommand{\ie}{\textit{i.e.}}

\title{Pano-NeRF: Synthesizing High Dynamic Range Novel Views with Geometry \\ from Sparse Low Dynamic Range Panoramic Images}

\author {
    Zhan Lu\textsuperscript{\rm 1},
    Qian Zheng\textsuperscript{\rm 2,\rm 3}\thanks{Corresponding author.},
    Boxin Shi\textsuperscript{\rm 4,\rm 5}, 
    Xudong Jiang\textsuperscript{\rm 1}
}

\affiliations {
    \textsuperscript{\rm 1} School of Electrical and Electronic Engineering, Nanyang Technological University \\
    \textsuperscript{\rm 2} College of Computer Science and Technology, Zhejiang University \\
    \textsuperscript{\rm 3} The State Key Lab of Brain-Machine Intelligence, Zhejiang University \\
    \textsuperscript{\rm 4} National Key Laboratory for Multimedia Information Processing, School of Computer Science, Peking University \\
    \textsuperscript{\rm 5} National Engineering Research Center of Visual Technology, School of Computer Science, Peking University \\
    zhan007@e.ntu.edu.sg, qianzheng@zju.edu.cn, shiboxin@pku.edu.cn, exdjiang@ntu.edu.sg
}

\begin{document}

\maketitle

\begin{abstract}
Panoramic imaging research on geometry recovery and High Dynamic Range (HDR) reconstruction becomes a trend with the development of Extended Reality (XR).
Neural Radiance Fields (NeRF) provide a promising scene representation for both tasks without requiring extensive prior data.
However, in the case of inputting sparse Low Dynamic Range (LDR) panoramic images, NeRF often degrades with under-constrained geometry and is unable to reconstruct HDR radiance from LDR inputs.
We observe that the radiance from each pixel in panoramic images can be modeled as both a signal to convey scene lighting information and a light source to illuminate other pixels.
Hence, we propose the irradiance fields from sparse LDR panoramic images, which increases the observation counts for faithful geometry recovery and leverages the irradiance-radiance attenuation for HDR reconstruction.
Extensive experiments demonstrate that the irradiance fields outperform state-of-the-art methods on both geometry recovery and HDR reconstruction and validate their effectiveness.
Furthermore, we show a promising byproduct of spatially-varying lighting estimation. The code is available at \url{https://github.com/Lu-Zhan/Pano-NeRF}.
\end{abstract}

\section{Introduction}
Panoramic imaging stands as a trend with the rise of extended reality (XR) for achieving immersive experiences, such as virtual walks in 360$^\circ$ scenes and inserting virtual objects with 360$^\circ$ lighting information.
These XR applications motivate panoramic imaging techniques, particularly in the tasks of geometry recovery~\cite{da20223d} and High Dynamic Range (HDR) reconstruction~\cite{yu2021luminance} under a panoramic scene.
Previous research yields promising outcomes via supervised learning on extensive pre-collected datasets or under specific predetermined conditions, \eg, a stereo camera system for geometry recovery~\cite{wang2020360sd} or controllable multi-exposure for HDR reconstruction~\cite{shen2011generalized}.
However, the efficacy of these approaches is profoundly influenced by the quality of the underlying training data and the imposed conditions.

Recently, Neural Radiance Fields (NeRF)~\cite{mildenhall2021nerf} emerged with a promising scene representation to recover geometry and radiance information through self-supervised training from multi-view images, which avoids extensive pre-collected data requirement.
Later research on sparse-view NeRF further improves the practicability of NeRF technique by reducing the number of multi-view images, where the recovered geometry is constrained by several priors, \eg, depth~\cite{niemeyer2022regnerf}, visibility~\cite{somraj2023vip}, and semantic feature~\cite{jain2021putting}.
Besides, several NeRF-based models exhibit the capability to reconstruct HDR radiance by requiring 
casual multi-exposed LDR images~\cite{huang2022hdr, jun2022hdr} or a pre-trained HDR reconstruction model~\cite{gera2022casual}.

\begin{figure}[t]
\centering
\includegraphics[width=\linewidth]{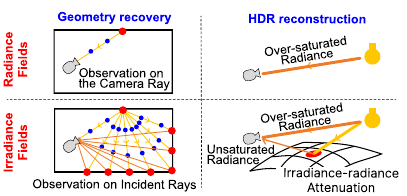}
\caption{Comparison between the existing radiance fields (top) and the proposed irradiance fields (bottom) on geometry recovery (left) and HDR reconstruction (right) from sparse LDR panoramic images. 
Radiance fields suffer from poor geometry recovery due to a few observation counts (number of blue dots) and cannot reconstruct HDR radiance with LDR inputs.
In contrast, the proposed irradiance fields recover faithful geometry by increasing the observation counts (number of blue dots) from incident rays (yellow lines) and infer over-saturated radiance from the unsaturated area by considering irradiance-radiance attenuation.}
\label{fig:irr_diff}
\end{figure}

However, NeRF-based models encounter two primary limitations: 1) existing sparse-view NeRF might fail to recover accurate geometry for a panoramic scene by using priors from objects rather than a panoramic scene.
And the scale variety of nearby/far objects in the panoramic scene further increases the trouble~\cite{barron2022mip};
2) they are unable to reconstruct HDR radiance from LDR inputs captured under a fixed exposure, due to the lack of a mechanism to address the ill-posed problem of HDR reconstruction.

Conversely, panoramic images exhibit a remarkable feature, where the radiance emitted by each pixel serves both as \textbf{a signal}, conveying scene lighting information to cameras through intrinsic factors (\eg, position, surface normal, and albedo) and as \textbf{a source light}, illuminating other pixels within the scene.
Based on this observation, this paper introduces irradiance fields to model the inter-reflection in panoramic scenes, via surface rendering~\cite{kajiya1986rendering}.
The irradiance fields consider the irradiance received from all incident light directions upon a given surface point and integrate them with the intrinsic factors of the surface point to yield the observed outgoing radiance.
Hence, the irradiance fields bring a distinct capability to
recover faithful geometry through the augmentation of the observation counts for volumetric particles with sparse inputs and reconstruct HDR radiance from multi-view LDR inputs by considering irradiance-radiance attenuation, shown in \Fref{fig:irr_diff}.
Furthermore, considering the feature of the radiance within panoramic images, 
the proposed irradiance fields could be within the existing radiance fields.
Therefore, we integrate our irradiance fields into radiance fields and perform joint optimization.
Our contributions are summarized as
\begin{itemize}
\item We propose the irradiance fields from sparse LDR panoramic images and explain how irradiance fields contribute to geometry recovery and HDR reconstruction.
\item We demonstrate the irradiance fields could be integrated
into and jointly optimized with radiance fields, based on
which we propose Pano-NeRF for geometry recovery and HDR novel views synthesis.
\item We show that Pano-NeRF achieves state-of-the-art performance regarding geometry recovery and HDR reconstruction. 
We further provide a byproduct of spatially varying lighting estimation.
\end{itemize}

\section{Related Work}
\label{related_work}
\noindent\textbf{Panoramic imaging.}
We introduce previous works on geometry recovery and HDR reconstruction dealing with panoramic images.
For geometry recovery, \cite{wang2020360sd} requires a stereo panoramic camera setup.
Several works estimate depth from the video recorded under free viewpoints~\cite{im2016all, zhang2020unsupervised}.
Recent works focus on the single-image panoramic geometry recovery via deep learning techniques and share the idea of the fusion of perspective depth estimation for a panoramic image~\cite{wang2020bifuse, yun2022improving, zhuang2022acdnet, ai2023hrdfuse}.
For HDR reconstruction, previous methods require multi-exposure LDR images for static poses~\cite{debevec1996modeling} or dynamic camera poses~\cite{chen2021hdr}.
Deep learning boosts the research on single-image HDR reconstruction on perspective images~\cite{li2019hdrnet, liu2020single, santos2020single} or panoramic images~\cite{yu2021luminance}).
Our method only requires sparse LDR panoramic images for geometry recovery and HDR reconstruction, to avoid extensive prior data or imposed conditions.

\noindent\textbf{Neural radiance fields.}
For geometry recovery by NeRF-based models, we only study sparse-view conditions.
Several works solve this problem by aggregating prior knowledge through pre-trained conditional radiance fields, \eg, deep features~\cite{yu2021pixelnerf, chibane2021stereo}, and 3D cost volume~\cite{chen2021mvsnerf}.
Other works constrain appearance consistency on the seen/unseen views to avoid pre-training~\cite{niemeyer2022regnerf, wynn2023diffusionerf}.
However, this work still recovers a blurry geometry since appearance regularizers cannot directly constrain the scene geometry.
Recent efforts on applying geometry priors would improve the recovered geometry, such as pre-estimated depth
(DS-NeRF~\cite{deng2022depth} and DD-NeRF~\cite{roessle2022dense}), depth smoothness (RegNeRF~\cite{niemeyer2022regnerf} and FreeNeRF~\cite{yang2023freenerf}), and geometry visibility (Ref-NeRF~\cite{verbin2022ref} and ViP-NeRF~\cite{somraj2023vip}).
360FusionNeRF~\cite{kulkarni2022360fusionnerf} requires a single RGBD panoramic image and trains NeRF by warping the RGBD inputs. 
IndoorPanoDepth~\cite{chang2023depth} builds a Signed Distance Field (SDF) and recovers geometry at the input views from sparse panoramic images by introducing a well geometry initialization.
For HDR reconstruction, Raw-NeRF~\cite{mildenhall2022nerf} proposes a model working with raw images for HDR radiance fields.
HDR-NeRF~\cite{huang2022hdr} builds HDR radiance fields with multi-exposure LDR inputs and paired exposure time.
HDR-Plenoxels~\cite{jun2022hdr} reduces the exposure time requirement and conducts self-calibrated for multi-exposure inputs.
PanoHDR-NeRF~\cite{gera2022casual} reconstructs HDR radiance from pre-estimated HDR panoramic images by an existing panoramic HDR reconstruction method.
The proposed irradiance fields help the geometry recovery and HDR reconstruction model by modeling inter-reflection within a panoramic scene, which is free of any pre-trained model, prior training data, or extra information.

\noindent\textbf{Implicit reflectance representation.}
Previous methods use reflectance fields to estimate intrinsic factors under conditions, \eg, known lighting~\cite{bi2020neural, srinivasan2021nerv}, or known geometry(\cite{yao2022neilf}).
Some methods distill the trained radiance fields to bake intrinsic factors and lighting conditions (NeRD~\cite{boss2021nerd} and NeRFactor~\cite{zhang2021nerfactor}), or indirect illumination~\cite{zhang2022modeling}.
Besides, TexIR~\cite{li2022multi} models the irradiance information with several HDR panoramic images (\ie, 14), pre-estimated mesh-based geometry, and semantic priors.
Our irradiance fields only require LDR inputs without any prior knowledge and exhibit the simplicity of sharing the scene representation with the same radiance fields.

\section{Irradiance Fields}
\label{sec:irr}

\subsection{Modeling Irradiance Fields}
\begin{figure}[t]
\centering
\includegraphics[width=\linewidth]{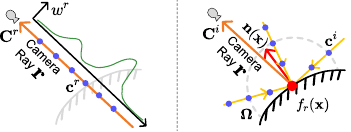}
% \vspace{-5pt}
\caption{An illustration of different outgoing radiance modeling between the radiance fields and the proposed irradiance fields. (Left) Radiance fields integrate the radiance $\mathbf{c}^r$ of each volumetric particle ({blue} dots) with weight $w^r$ along the camera ray $\mathbf{r}$ (original line). (Right) Irradiance fields integrate the radiance $\mathbf{c}^i$  from the incident light directions (yellow lines) with weight $f_r(\mathbf{x}, \boldsymbol{\omega}_o, \boldsymbol{\omega}_i) \cdot \boldsymbol{\omega}_i \mathbf{n}^\top$.}
% \vspace{-5pt}
\label{fig:irr_model}
\end{figure}

\noindent\textbf{Preliminary of radiance fields.}
The radiance fields in NeRF~\cite{mildenhall2021nerf} assume that the scene is composed of a cloud of light-emitting particles and the volumetric particles have emission and transmittance to pass through the radiance from other volumetric particles~\cite{tagliasacchi2022volume}, as shown in \Fref{fig:irr_model}~(a).
The outgoing radiance $\mathbf{C}^r$ of position $\mathbf{x}$ is computed through volume rendering, \ie, integrate radiance $\mathbf{c}^r$ of different volumetric particles (weighted by $w^r$) distributed along the camera ray $\mathbf{r}$ between viewpoint $\mathbf{o}$ and $\mathbf{x}$.
Due to the unavailability of $\mathbf{x}$, $\mathbf{r}$ is formulated as $\mathbf{r}(t) = \mathbf{o} + t \boldsymbol{\omega}_o, t\in[t_n, t_f]$  with view direction $\boldsymbol{\omega}_o$, and near/far bounds $t_n$/$t_f$.
The calculation of outgoing radiance $\mathbf{C}^r$ is formulated as\footnote{We provide a more comprehensive formulation showing the relationship between different variables according to the implementation in~\cite{mildenhall2021nerf}.},
\begin{equation}
\footnotesize
\label{eqn:volume}
\begin{aligned}
	&\mathbf{C}^r(\mathbf{x}) = \mathbf{C}^r(\mathbf{r}) = \int^{t_f}_{t_n} w^r(\mathbf{o}, \boldsymbol{\omega}_o, t) \cdot \mathbf{c}^r(\mathbf{o}, \boldsymbol{\omega}_o, t) d t, \\
	&\text{s.t.} w^r(\mathbf{o}, \boldsymbol{\omega}_o, t)= \exp(-\int^{t}_{t_n} \sigma(\mathbf{o}, \boldsymbol{\omega}_o, t) d t) \cdot \sigma(\mathbf{o}, \boldsymbol{\omega}_o, t),
\end{aligned}
\end{equation}
where $\mathbf{c}^r$ represents the radiance color of each volumetric particle that determines emission, and $w^r$ is the weight calculated based on the density $\sigma$ of volumetric particles.
The radiance fields in NeRF~\cite{mildenhall2021nerf} are implemented based on a multilayer perception (MLP) network, 
\begin{equation}
\label{eqn:ra_func}
\begin{aligned}
\text{MLP}(\mathbf{r}) \rightarrow (\mathbf{c}^r, \sigma),
\end{aligned}
\end{equation}

\noindent\textbf{Irradiance fields formulation.}
 Our irradiance fields assume that the scene is composed of several surface pieces with reflection effect only and the radiance is the interaction result of incoming light, reflectance, and surface normal, as shown in \Fref{fig:irr_model}~(b).
Different from the radiance fields, our irradiance fields compute the outgoing radiance of position $\mathbf{x}$ through surface rendering~\cite{kajiya1986rendering}, \ie, integrate irradiance $\boldsymbol{\omega}_i \mathbf{n}^\top \cdot \mathbf{c}^i$ around different incoming light directions $\boldsymbol{\omega}_i$ distributed on sphere $\boldsymbol{\Omega}$ centered at $\mathbf{x}$,
\begin{equation}
\footnotesize
\label{eqn:irr_model}
\begin{aligned}
&\mathbf{C}^i(\mathbf{x}) =  \int_{\boldsymbol{\omega}_i\in\boldsymbol{\Omega}} f_r(\mathbf{x}, \boldsymbol{\omega}_o, \boldsymbol{\omega}_i) \cdot \boldsymbol{\omega}_i \mathbf{n}^\top \cdot \mathbf{c}^i(\mathbf{x}, \boldsymbol{\omega}_i) d\boldsymbol{\omega}_i, \\
\end{aligned}
\end{equation}
where $\mathbf{c}^i$ indicates the radiance from each incoming light direction  and $f_r$ is the bidirectional reflection distribution function (BRDF), $\mathbf{n}$ is the surface normal at point $\mathbf{x}$.
Note $\boldsymbol{\omega}_i \mathbf{n}^\top$ is replaced as $\text{max}(\boldsymbol{\omega}_i \mathbf{n}^\top, 0)$ in the implementation.

\noindent\textbf{Geometry recovery with sparse inputs.}
As illustrated in \Fref{fig:irr_diff} (left-top), sparse inputs bring a small number of observation counts, which might cause unconstrained geometry~\cite{niemeyer2022regnerf}.
As can be observed from \Fref{fig:irr_diff} (left-bottom), the irradiance fields generate several rays along the incident light direction $\boldsymbol{\omega}_i$ for each position $\mathbf{x}$.
The $\mathbf{c}^i$ is equal to the radiance $\mathbf{C}^{r}$ of each incident ray that could be calculated in~\Eref{eqn:volume}.
Thus, the proposed irradiance fields are expected to increase the observation counts for volumetric particles (\Fref{fig:irr_diff} (left)) by considering the inter-reflection between irradiance and outgoing radiance.
Although $\mathbf{c}^i$ is not directly supervised, its integral is constrained through the training error between the observed radiance $\mathbf{C}_\text{gt}$ and $\mathbf{C}^i$, hence facilitates the optimization of $\sigma$ and $\mathbf{c}^r$.
Therefore, the proposed irradiance fields could achieve faithful geometry recovery even with sparse inputs.

\noindent\textbf{HDR reconstruction with LDR inputs.}
The HDR radiance could be well restored from LDR one if it is unsaturated (\eg, applying inverse tone mapping~\cite{rempel2007ldr2hdr}).
The primary difficulty of HDR reconstruction is from over-saturated regions, where the clipping operation drops vital information. 
Fortunately, over-saturated regions do not always cover the whole panoramic image (\eg, most are light sources in the indoor scene), and the proposed irradiance fields could leverage the radiance in unsaturated regions to reconstruct HDR radiance.
As illustrated in \Fref{fig:irr_diff} (right), the basic idea is to back-propagate the training error (from points in unsaturated regions) between the observed radiance $\mathbf{C}_\text{gt}$ (from input images) and $\mathbf{C}^i$ to optimize HDR $\mathbf{c}^i$ in over-saturated regions, based on \Eref{eqn:irr_model}.
This can be achieved due to the attenuation effect brought by the BRDF (\ie, $\phi<1$) and the cosine of the incident angle (\ie, $\boldsymbol{\omega}_i\mathbf{n}^\top<1$).
Additionally, $\mathbf{c}^i$ in over-saturated regions could be further optimized based on the observed radiance $\mathbf{C}^r$ according to \Eref{eqn:cc}.

\begin{figure*}[t]
\centering
\includegraphics[width=\linewidth]{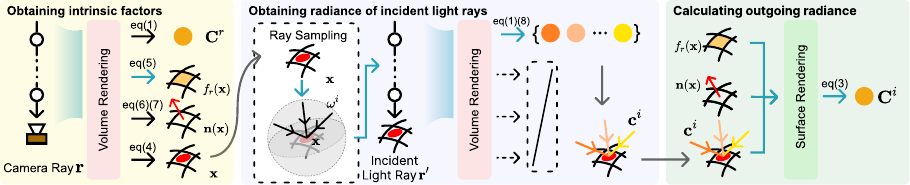}
\caption{Illustration of calculating the outgoing radiance $\mathbf{C}^{i}$ from the conventional radiance fields.
Firstly, we obtain the intrinsic factors on a surface point by integrating the BRDF output $\Phi$, derived density $d\sigma$, and distance $t$ along the camera ray $\mathbf{r}$ via volume rendering, as same as the computation of radiance $\mathbf{C}^{r}$.
Secondly, We sample the incident light rays $\mathbf{r}^\prime$ on the surface point $\mathbf{x}$ from directions distributed uniformly on the sphere centered at $\mathbf{x}$, and then compute the radiance $\mathbf{c}^{i}$ of sampled incident light rays.
At last, we integrate obtained BRDF $f_r(\mathbf{x})$, surface normal $\mathbf{n}(\mathbf{x})$, and incident light rays (with $\mathbf{c}^{i}$ and direction $\omega^{i}$) to calculate outgoing radiance $\mathbf{C}^{i}$.}
% \vspace{-5pt}
\label{fig:framework}
\end{figure*}

\subsection{Optimizing Irradiance Fields}
\label{subsec:optimization}
This section introduces how to integrate our irradiance fields into the radiance fields and perform joint optimization.
Specifically, we focus on the calculation of outgoing radiance $\mathbf{C}^i$ based in \Eref{eqn:irr_model}.

\noindent\textbf{Obtaining intrinsic factors.}
We follow a numerical solution~\cite{zhang2021nerfactor} sharing a similar idea as volume rendering~\cite{kajiya1984ray} in \Eref{eqn:volume} to calculate the position of surface point $\mathbf{x}$ by
\begin{equation}
\footnotesize
\label{eqn:distance}
\begin{aligned}
	\mathbf{x} = \mathbf{o} + (\int^{t_f}_{t_n} w^r(\mathbf{o}, \boldsymbol{\omega}_o, t) \cdot t d t)\boldsymbol{\omega}_o,
\end{aligned}
\end{equation}
where $w^r(\mathbf{o}, \boldsymbol{\omega}_o, t)$ is same as that in \Eref{eqn:volume}.

As suggested by many scene understanding works (\eg, \cite{xing2018automatic, zhou2019glosh, wang2021learning}), we assume the BRDF in irradiance fields to be diffuse (Lambertain model), \ie, $f_r$ only depends on the position of 3D point $\mathbf{x}$ while being free from the view direction,
\begin{equation}
\label{eqn:brdf}
\begin{aligned}
	f_r(\mathbf{x}) = f_r(\mathbf{r}) = \int^{t_f}_{t_n} w^r(\mathbf{o}, \boldsymbol{\omega}_o, t) \cdot \phi(\mathbf{o}, \boldsymbol{\omega}_o, t) d t,
\end{aligned}
\end{equation}
where $\phi_r(\mathbf{o}, \boldsymbol{\omega}_o, t)$ is the albedo for a volumetric particle.
Though the diffuse assumption limits the expression of non-Lambertain surfaces, it affects less since 1) non-Lambertian surfaces occupy less area than the Lambertian ones in practice and 2) we leverage the outputs from radiance fields (non-Lambertain model) as the reconstructed results.

According to~\cite{verbin2022ref}, the density $\sigma(\mathbf{o}, \boldsymbol{\omega}_o, t)$ for each volumetric particle can be leveraged to calculate its surface normal $\mathbf{n}(\mathbf{o}, \boldsymbol{\omega}_o, t)$, 
\begin{equation}
\label{eqn:normal}
\begin{aligned}
     & \mathbf{n}(\mathbf{o}, \boldsymbol{\omega}_o, t)=-\frac{d \sigma(\mathbf{o}, \boldsymbol{\omega}_o, t)}{\|d \sigma(\mathbf{o}, \boldsymbol{\omega}_o, t)\|},
\end{aligned}
\end{equation}
and the surface normal for position $\mathbf{x}$ can be numerically calculated by 
\begin{equation}
\label{eqn:normal}
\begin{aligned}
    &\mathbf{n}(\mathbf{x}) = \int^{t_f}_{t_n} w^r(\mathbf{o}, \boldsymbol{\omega}_o, t) \cdot \mathbf{n}(\mathbf{o}, \boldsymbol{\omega}_o, t) d t, \\
\end{aligned}
\end{equation}
Note that each surface normal is normalized to be a unit vector before further calculation.

\noindent\textbf{Obtain radiance of incident light rays.}
As shown in \Fref{fig:framework}, we first sample the incident light directions $\boldsymbol{\omega}_i$ on the surface point $\mathbf{x}$ to determine the incident light rays $\mathbf{r}^\prime(t) = \mathbf{x} + t\boldsymbol{\omega}_i$, then obtain the radiance $\mathbf{c}^i(\mathbf{x}, \boldsymbol{\omega}_i)$,
\begin{equation}
\begin{aligned}
\label{eqn:cc}
    &\mathbf{c}^i(\mathbf{x}, \boldsymbol{\omega}_i) = \mathbf{C}^r(\mathbf{r}^\prime),   
\end{aligned}
\end{equation}

\noindent\textbf{Calculating outgoing radiance.}
The outgoing radiance $\mathbf{C}^i(\mathbf{x})$ is then calculated with the estimated BRDF $f_r$, surface normal $\mathbf{n}(\mathbf{x})$ and a grouped radiance $\{\mathbf{c}^i\}$ of incident light rays via surface rendering in~\Eref{eqn:irr_model}.

In summary, the proposed irradiance fields can be jointly optimized with radiance fields since all variables except $\phi$ can be directly calculated from radiance fields.
We modify the MLP in a NeRF-based method to output an additional variable $\phi$, and \Eref{eqn:ra_func} is revised as
\begin{equation}
\begin{aligned}
\text{MLP}(\mathbf{r}) \rightarrow (\mathbf{c}^r, \sigma, \phi),
\end{aligned}
\end{equation}

\subsection{Pano-NeRF}
The overview of irradiance fields is displayed in \Fref{fig:framework}.

\noindent\textbf{Network structure.}
We build our irradiance fields upon the model of Mip-NeRF~\cite{barron2021mip} and modify it in the following aspects.
First, we increase the output range of $\mathbf{c}^r$ to consider HDR radiance, by changing the output activation function of ReLU to Softplus.
Second, we add output $\phi\in\mathbf{R}^3$ to the MLP for albedo estimation.

\noindent\textbf{Geometry prior.}
The recovered geometry in the radiance fields often reveals a thick surface (`foggy' even with dense inputs~\cite{verbin2022ref}) and rough surface ~\cite{niemeyer2022regnerf}.
Such geometry might limit the reconstruction of irradiance fields as the irradiance fields require a rough geometry for ray sampling and surface rendering computation.
Hence we adopt the geometry prior $\mathcal{R}_{v}$ indicated in~\cite{verbin2022ref} to produce thin and smooth surface, that is,
\begin{equation}
\label{eqn:nor_reg}
\footnotesize
\begin{aligned}
	\mathcal{R}_{v} = 
    &\sum_{\mathbf{o}, \boldsymbol{\omega}_o} (\int_{t_n}^{t_f} w^r(\mathbf{o}, \boldsymbol{\omega}_o, t) \cdot \mathrm{max}(-\boldsymbol{\omega}_o\mathbf{n}^\top(\mathbf{o}, \boldsymbol{\omega}_o, t), 0)^{2} dt),
\end{aligned}
\end{equation}
The minimization of $\mathcal{R}_{v}$ ensures a thin and smooth surface by preventing the model from expressing too many visible volumetric particles around the surface.

\begin{table*}[t]
\centering
% \normalsize
\small
% \singlespacing
% \resizebox{0.7\linewidth}{!}{
\begin{tabular}{lcccccccc}
\toprule
\multirow{3}{*}{Methods}  & \multicolumn{5}{c}{Synthetic} & \multicolumn{3}{c}{Real} \\ \cmidrule(lr){2-6} \cmidrule(lr){7-9} 
    & Depth & Normal  & \multicolumn{3}{c}{Image}  & \multicolumn{3}{c}{Image}  \\ \cmidrule(lr){2-2} \cmidrule(lr){3-3} \cmidrule(lr){4-6} \cmidrule(lr){7-9}  
    & RMSE$\downarrow$ & MAE$\downarrow$ & PSNR$\uparrow$ & SSIM$\uparrow$   & LPIPS$\downarrow$ & PSNR$\uparrow$ & SSIM$\uparrow$   & LPIPS$\downarrow$ \\ \midrule
Omnifusion & 1.13 & - & - & - & - & - & - & - \\ 
IndoorPanoDepth  & 1.29 & 45.35 & 18.57  & 0.61 & 0.43 & 15.91  & 0.59 & 0.47 \\ 
Mip-NeRF & 1.06 & 65.44 & 19.91  & 0.66 & 0.41 & 16.77  & 0.61 & 0.45  \\  
Mip-NeRF+$\mathcal{R}_v$    & 1.05	& \underline{36.81}   & 20.54  & 0.69 & 0.40 & 17.23  & 0.62 & 0.42 \\
RegNeRF  & \underline{0.77} & 63.98 & \underline{22.22}  & \underline{0.74} & 0.34 & \underline{19.88}  & \underline{0.68} & 0.36 \\
FreeNeRF     & 1.88       & 66.76        & 21.23  & 0.68 & \underline{0.32} & 19.23  & 0.67 & \underline{0.35}    \\
Ours          & \textbf{0.72}     & \textbf{29.03}  & \textbf{23.10}  & \textbf{0.78} & \textbf{0.31}    & \textbf{21.08}  & \textbf{0.76} & \textbf{0.33}   \\ \bottomrule
\end{tabular}%}
\caption{Comparison of geometry recovery and novel view synthesis.}
\label{tab:nvs}
\end{table*}

\noindent\textbf{Albedo priors.}
As we only leverage the observed pixel color for optimization, the freedom of the new variable $\phi$ might cause the radiance color distortion as the neural network is more easily to express color variety by $\phi$ rather than the irradiance.
To mitigate the impact from $\phi$, we take two priors to constrain the albedo estimation of our model.
First, we limit the range of the estimated $\phi$ with $[0.03, 0.8]$ as suggested in~\cite{ward1998rendering}.
Second, we encourage the chromaticity of the estimated $f_r$ to be similar to that of corresponding observed radiance $\mathbf{C}_\text{gt}$,
\begin{equation}
\footnotesize
\begin{aligned}
	\label{eqn:chrom_reg}
	\mathcal{R}_{c} &=\sum_{\mathbf{x}\in\boldsymbol{\Phi}}{\left\|\frac{f_r(\mathbf{x})}{\|f_r(\mathbf{x})\|}-\frac{\mathbf{C}_\text{gt}(\mathbf{x})}{\|\mathbf{C}_\text{gt}(\mathbf{x})\|}\right\|^2_{2}},
\end{aligned}
\end{equation}
where set $\boldsymbol{\Phi}$ contains positions of all surface points in a scene.
Such constraints of the albedo are able to maintain the chromaticity of the reconstructed radiance color, which is enough to produce pleasurable results.

\noindent\textbf{Loss function.}
We adopt the same loss function from Mip-NeRF to constrain the estimated radiance via volume rendering.
To convert HDR radiance output to LDR value, we apply empirical tone-mapping~\cite{arrighetti2017academy, chen2022text}, clipping (into $[0, 1]$), and gamma correction (factor of $2.2$) on the output radiance to obtain mapped LDR value.
MSE loss $\mathcal{L}_{r}$ for the radiance fields is calculated in coarse and fine phases as,
\begin{equation}
\label{eqn:ra_mseloss}
% \footnotesize
\begin{aligned}
	\mathcal{L}_{r}=\sum_{\mathbf{o}, \boldsymbol{\omega}_o}(\alpha_{1}&\left\|g(\mathbf{C}^r_{c}(\mathbf{o}, \boldsymbol{\omega}_o))-\mathbf{C}_{\mathrm{gt}}(\mathbf{o}, \boldsymbol{\omega}_o)\right\|^2_{2} + \\
	&\left\|g(\mathbf{C}^r_{f}(\mathbf{o}, \boldsymbol{\omega}_o))-\mathbf{C}_{\mathrm{gt}}(\mathbf{o}, \boldsymbol{\omega}_o)\right\|^2_{2}),
\end{aligned}
\end{equation}
where $\mathbf{C}^r_{c}$ and $\mathbf{C}^r_{f}$ are the estimated radiance in coarse and fine phases, respectively. 
$g(\cdot)$ represents the process of HDR to LDR conversion. $\alpha_1$ is set to 0.1 as in Mip-NeRF.
The MSE loss $\mathcal{L}_{i}$ for the irradiance fields is calculated only in the fine phase as,
\begin{equation}
	\label{eqn:irr_mseloss}
	\mathcal{L}_{i}=\sum_{\mathbf{o}, \boldsymbol{\omega}_o} \left\|g(\mathbf{C}^{i}(\mathbf{o}, \boldsymbol{\omega}_o))-\mathbf{C}_{\mathrm{gt}}(\mathbf{o}, \boldsymbol{\omega}_o)\right\|^2_{2},
\end{equation}
Therefore, the overall loss function is,
% calculated with the geometry prior $\mathcal{R}_{v}$, and the albedo prior $\mathcal{R}_{c}$,
\begin{equation}
	\label{eqn:loss_function}
	\mathcal{L}= \mathcal{L}_{r} + \alpha_{2} \mathcal{L}_{i} + \alpha_{3} \mathcal{R}_{v} + \alpha_{4} \mathcal{R}_{c},
\end{equation}
We empirically set $\alpha_{2} = 1$, $\alpha_{3} = 0.1$, and $\alpha_{4} = 1$ to balance loss scale and stabilize the model training.

\section{Experiments}
\subsection{Setup}
\noindent\textbf{Synthetic data.} We render evaluation data from 5 3D scene models, including `classroom', `barbershop', `living room', `bedroom', and `gallery'.
Besides, we collect the data from the 3D scanned HDR dataset Replica~\cite{replica19arxiv} and select 8 enclosed scenes, including `apartment', `hotel', `bathroom', `office-0', `office-1', `office-4', `room-0', and `room-1'.
For each scene, we sample 100 camera poses and render the LDR panoramic image and the paired ground truths (\ie, HDR panoramic image, depth maps, and surface normal map) at each sampled camera pose.

\noindent\textbf{Real captured data.}
We capture 30 HDR panoramic images in 3 real scenes, including `real-meeting room', `real-bedroom', and `real-classroom', representing the large, medium, and small indoor spaces.
Besides, we scan each scene with the LiDAR camera in iPhone 13 Pro to obtain the depth and surface normal for reference.
We implement an open-source tool openMVG to calibrate camera poses and merge HDR panoramic images with 9 bracketed exposures. 

\begin{figure*}[t]
\centering
\includegraphics[width=\linewidth]{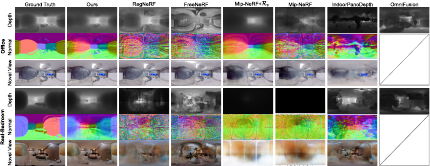} 
\caption{Comparison of geometry recovery in term of depth maps (1st\&4th rows), surface normal maps (2nd\&5th rows), and panoramic images (3rd\&6th rows). The ground truth of depth and surface normal for the real scene is only for visual reference.}
% \vspace{-5pt}
\label{fig:nvs}
\end{figure*}

\noindent\textbf{Evaluation settings}
% To evaluate irradiance fields with sparse inputs, for each time training, we randomly select 3 LDR panoramic images to train models as the worst case in~\cite{niemeyer2022regnerf}.
For each scene, we randomly select 3 LDR panoramic images to train models, and the rest are used for evaluation.
We repeat this procedure 10 times and compute the mean performance to alleviate the randomness of the results.
The resolution of images is set to 128 $\times$ 256.

\noindent\textbf{Training details} 
The sample numbers of volumetric particles along a camera ray and an incoming lighting ray are 64 and 10, respectively.
The sample number of directions for incoming lighting rays towards each surface point $\mathbf{x}$ is 80.
We train Pano-NeRF with the following settings: 44k optimization iterations, using Adam~\cite{kingma2014adam} optimizer with hyper-parameters $\beta_{1}=0.9$, $\beta_{2}=0.999$, $\epsilon=10^{-6}$, a log-linearly annealed learning rate from $2 \times 10^{-4}$ to $2 \times 10^{-5}$ with a warm-up phase of 2500 iterations.
Our model and ablations take 512 as the batch size to keep a close number of sampled volumetric particles per iteration with Mip-NeRF~\cite{barron2021mip}, whose batch size is 4096.
We optimize the only radiance fields at the first 8.8k iterations and then perform joint optimization on two fields.

\subsection{Overall Performance}

\noindent\textbf{Geometry Recovery from sparse inputs.}
We compare our method with two panoramic depth estimation methods, including Omnifusion~\cite{li2022omnifusion} and IndoorPanoDepth~\cite{chang2023depth}.
Omnifusion is a deep learning-based method trained with large-scale data, 
while IndoorPanoDepth is an SDF-based method using the same sparse panoramic inputs.
We additionally add the two state-of-the-art sparse-view NeRF-based methods, \ie, RegNeRF~\cite{niemeyer2022regnerf}, FreeNeRF~\cite{yang2023freenerf}.
% We additionally add the two state-of-the-art sparse-view NeRF-based methods, \ie, RegNeRF\footnote{RegNeRF is trained only with geometry regularization while appearance regularization is not provided in their released codes.}~\cite{niemeyer2022regnerf}, 
% \footnote{The frequency attenuation is used for facing forward scenes as suggested in FreeNeRF.}
Besides, we take Mip-NeRF~\cite{barron2021mip} and its ablated version Mip-NeRF+$\mathcal{R}_v$ as the baseline methods.
We report the geometry metrics, and LDR image quality metrics (including PSNR, SSIM~\cite{wang2004image}, and LPIPS~\cite{zhang2018perceptual}).
Geometry metrics are not calculated for real scenes due to no aligned ground truth.

\begin{figure*}[t]
\centering
\includegraphics[width=0.8\linewidth]{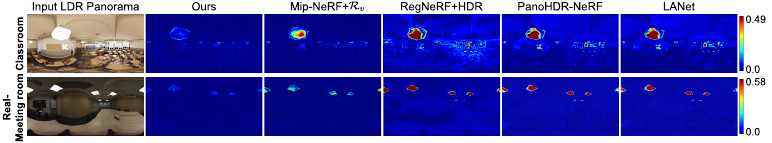} 
\caption{Comparison of HDR reconstruction on input views. The first column is the input LDR panoramic image, and 2nd to 5th columns show the error maps between the reconstructed HDR panoramic images and the ground truth. }
% \vspace{-5pt}
\label{fig:hdr}
\end{figure*}
\begin{table*}[t]
\centering
\small
\begin{tabular}{lcccc}
\toprule
Methods & PU-PSNR$\uparrow$ & PU-SSIM$\uparrow$ & HDR-VDP3$\uparrow$ & RMSE$\downarrow$ \\ \midrule
LANet       & 31.9274	& 0.9317 & 7.7622 & 0.9195 \\
PanoHDR-NeRF & 33.0560	& 0.9415 & 7.9560 & 0.7343 \\
RegNeRF+HDR & 31.3735	& 0.8638 & 7.3694 & 0.9064 \\
Mip-NeRF+$\mathcal{R}_v$ & \underline{41.6452}	& \underline{0.9875}	& \underline{9.6121} & \underline{0.3110} \\
Ours& \textbf{43.4662}	& \textbf{0.9899}	& \textbf{9.6745} & \textbf{0.2686} \\ \bottomrule
\end{tabular}%}%}
\caption{Comparison of HDR reconstruction on input views.}
\label{tab:hdr}
\end{table*}

As shown in \Tref{tab:nvs}, Pano-NeRF outperforms others on the results of geometry recovery.
\Fref{fig:nvs} demonstrates the qualitative comparison of geometry recovery, where we have the same conclusion that Pano-NeRF could achieve more faithful depth and accurate surface normal.
Besides, Pano-NeRF provides the best performance on novel view synthesis, as shown in \Tref{tab:nvs} and \Fref{fig:nvs} (estimating a clear panoramic image with fewer artifacts.).
The reason could be that a well-recovered geometry is crucial to novel view synthesis in the case of sparse inputs.
Furthermore, by comparing the Mip-NeRF and Mip-NeRF+$\mathcal{R}_v$, we observe that directly applying geometry prior $\mathcal{R}_v$ could only benefit a smoother surface but no improvement on depth estimation or cause the severe degradation on novel views (shown in \Fref{fig:nvs}).
In contrast, the irradiance fields not only benefit from the smooth surface for better geometry but also prevent the collapse of geometry recovery.
% These results validate the proposed irradiance fields do help to improve the geometry recovery from sparse inputs.

\noindent\textbf{HDR Reconstruction from LDR inputs}.
We compare our method with the state-of-the-art panoramic HDR reconstruction method LANet~\cite{yu2021luminance} and three NeRF-based methods, including PanoHDR-NeRF~\cite{gera2022casual}, RegNeRF~\cite{niemeyer2022regnerf}, and our baseline Mip-NeRF~\cite{barron2021mip} with $\mathcal{R}_{v}$.
Since RegNeRF is not implemented for the HDR reconstruction, we take pre-estimated HDR panoramic images from LANet for its training, namely `RegNeRF+HDR'.
In this experiment, only the reconstructed HDR panoramic images at input viewpoints are evaluated on each method to keep the same view requirement of LANet.
We adopt commonly used HDR metrics for HDR image evaluation, including PU-PSNR~\cite{azimi2021pu21}, PU-SSIM~\cite{azimi2021pu21}, HDR-VDP3~\cite{mantiuk2023hdr}, and RMSE.

\Tref{tab:hdr} demonstrates the quantitative comparison.
LANet reports limited performance of HDR reconstruction, as they might suffer from poor generalization from their training data to testing scenarios. 
NeRF-based methods PanoHDR-NeRF and RegNeRF+HDR keep a close performance with LANet since they could only learn HDR information from the noisy HDR prediction.
Besides, RegNeRF+HDR drops significantly in terms of PU-SSIM, which indicates that the inconsistent HDR reconstruction of multi-views might cause degradation in the novel view synthesis.
Mip-NeRF+$\mathcal{R}_{v}$ demonstrates better results as it takes advantage of the tone mapping operation.
In contrast, Pano-NeRF achieves the best performance on HDR reconstruction even compared with the baseline.
Observed from in \Fref{fig:hdr}, we find that Pano-NeRF accurately predicts the light source where often over-saturated as the HDR reconstruction results are close to the ground truth.
The reason could be that our irradiance fields constrain these over-saturated pixels by the information learned from the unsaturated area.

\begin{figure*}[t]
\centering
\includegraphics[width=1.0\linewidth]{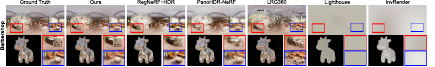}
\caption{Comparison of HDR reconstruction on novel views and inserted mirror-like objects. Zoom in for better details.}
% \vspace{-5pt}
\label{fig:sv_lit}
\end{figure*}

\subsection{Validation of Irradiance Fields}
To validate the effectiveness of the proposed irradiance fields, we take the Mip-NeRF~\cite{barron2021mip} as our baseline method representing radiance fields.
Then, we simply integrate the proposed irradiance fields into the baseline, namely `Mip-NeRF+Irr', as the counterpart.
Besides, we study the impact of the geometry prior $\mathcal{R}_{v}$ as it could be applied to both methods, namely `Mip-NeRF+$\mathcal{R}_{v}$' and `Mip-NeRF+Irr+$\mathcal{R}_{v}$', respectively.
We report linear Root Mean Square Error (RMSE) for the estimated depth, Mean Angle Error (MAE) (in degree) for the estimated surface normal (`normal' for short), and HDR-VDP3~\cite{mantiuk2023hdr} for HDR novel view synthesis.
% Note we conduct this experiment only on 13 synthetic scenes with the ground truth of geometry.

\begin{figure}[t]
\centering
\includegraphics[width=\linewidth]{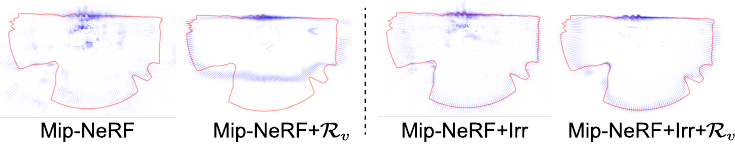}    
\caption{Comparison of the recovered geometry observed from an
overhead angle. The blue point represents weight $\omega^{r}$ of the volumetric particle at that position (The brighter indicates a larger $\omega^{r}$). The red line is the ground truth.}
\label{fig:pc}
\end{figure}

Observed from \Tref{tab:ablation}, irradiance fields benefit the performance on both tasks.
Specifically, irradiance fields improve significantly in the depth estimation and HDR novel view quality by simply adding to the radiance fields and performing joint optimization.
Besides, we find that $\mathcal{R}_{v}$ could further boost the performance of irradiance fields, in contrast, it only contributes to the better surface normal when applying to the radiance fields. 
We demonstrate an example of reconstructed geometry in \Fref{fig:pc}. 
We found that with the help of irradiance fields, the reconstructed geometry would be more accurate to the ground truth than that in radiance fields (Mip-NeRF) from sparse inputs.
Besides, $\mathcal{R}_{v}$ could lead to a smoother surface but might significantly degrade the depth estimation, which is not suitable for novel view synthesis with an unconstrained depth.
In contrast, the irradiance fields benefit from the smooth surface from  $\mathcal{R}_{v}$ and predict a more faithful depth.

\begin{table}[t]
\center 
\small
\begin{tabular}{lccc}
\toprule
\multirow{1}{*}{Methods}  & HDR-VDP3$\uparrow$   & D-RMSE$\downarrow$ & N-MAE$\downarrow$  \\ \midrule
Mip-NeRF & 7.1165 & 1.0576	& 65.4398  	\\
Mip-NeRF+$\mathcal{R}_{v}$ & 7.2328 & 1.0524	& \underline{36.8086}   	\\ \midrule
Mip-NeRF+Irr & \underline{7.5852}  & \underline{0.8565}	& 56.6353	\\
Mip-NeRF+Irr+$\mathcal{R}_{v}$  & \textbf{7.6693} & \textbf{0.8146} & \textbf{30.4295} 	\\ \bottomrule
\end{tabular}%}
\caption{Validation of the proposed irradiance fields.}
\label{tab:ablation}
\end{table}

\subsection{Spatially-varying Lighting Estimation}
We demonstrate that Pano-NeRF could be further used for spatially varying lighting estimation, as a byproduct of HDR novel view synthesis.
We compare with NeRF-based methods and additional spatially-varying lighting estimation methods, including InvIndoor~\cite{li2020inverse}, Lighthouse~\cite{srinivasan2020lighthouse}, and LRG360~\cite{li2021lighting}.
% \footnote{Implementation is at \url{https://github.com/Lu-Zhan/Pano-NeRF}.}.
\Fref{fig:sv_lit} shows that Lighthouse and InvRender produce low-frequency lighting maps, and LRG360 brings the artifacts into the lighting maps.
RegNeRF+HDR and PanoHDR-NeRF keep similar results and suffer from inconsistent HDR reconstruction for different views to produce blurry lighting maps.
Conversely, Pano-NeRF outperforms others as it provides spatially-varying HDR lighting maps with the richest details close to the ground truth, which benefits more realistic mirror-like objects inserted with accurate specular.

\section{Conclusion}
In summary, this paper introduces irradiance fields as a novel solution for geometry recovery and HDR reconstruction with sparse LDR panoramic images. 
The irradiance fields consider the inter-reflection in panoramic scenes, to recover faithful geometry from sparse inputs by increasing observation counts of volumetric particles and to reconstruct accurate HDR radiance from LDR inputs by yielding irradiance-radiance attenuation.
We integrate the irradiance fields into the scene representation of radiance fields and perform joint optimization.
Extensive experiments validate the effectiveness of irradiance fields and demonstrate superior performance on geometry recovery and HDR reconstruction.
We further investigate the potential of the proposed irradiance fields for estimating spatially varying lighting.

% \noindent\textbf{Limitation.}
% Since our method optimizes two fields (\eg, radiance and irradiance) but the baseline method Mip-NeRF only considers one (\eg, radiance), our method needs about doubled the optimization time under the same settings.

\section{Acknowledgments}
This research is supported in part by the Centre for Information Sciences and Systems (CISS) of School of Electrical \& Electronic Engineering and CARTIN at Nanyang Technological University, in part by the State Key Lab of Brain-Machine Intelligence at Zhejiang University, and in part by the National Natural Science Foundation of China (Grant Nos. 62088102 and 62136001).

\bibliography{camera_ready.bib}

\end{document}